# CROSS-TALK BASED MULTI-TASK LEARNING FOR FAULT CLASSIFICATION OF MACHINE SYSTEM INFLUENCED BY MULTIPLE VARIABLES

Wonjun Yi, Rismaya Kumar Mishra, Yong-Hwa Park

*Korea Adavanced Institute of Science and Technology (KAIST), Republic of Korea*
*email: lasscap@kaist.ac.kr*

Machine systems inherently generate signals in which fault conditions and various variables influence signals measured from machine system. Although many existing fault classification studies rely solely on direct fault labels, the aforementioned signals naturally embed additional information shaped by other variables. Herein, we leverage this through a multi-task learning (MTL) framework that jointly learns fault conditions and other variables influencing measured signals. Among MTL architectures, cross-talk structures have distinct advantages because they allow for controlled information exchange between tasks through the cross-talk layer while preventing negative transfer, in contrast to shared trunk architectures that often mix incompatible features. We build on our previously introduced residual neural dimension reductor model, and extend its application to two benchmarks where system influenced by multiple variables. The first benchmark is a drone fault dataset, in which machine type and maneuvering direction significantly alter the frequency components of measured signals even under the same drone status. The second benchmark dataset is motor compound fault dataset. In this system, severity of each fault component, inner race fault, outer race fault, misalignment, and unbalance influences measured signal. Across both benchmarks, our residual neural dimension reductor, consistently outperformed single-task models, multi-class models that merge all label combinations, and shared trunk multi-task models.

---

## 1. Introduction

With the growing needs of condition monitoring and predictive maintenance, fault classification is becoming increasingly important. For machine system targeted for fault classification, measured signal is influenced by multiple variables including severities and other variables such as operation condition and machine type. Threrfore, in inverse problem point of view, classification task of each fault or variables influencing signal measured from machine system can be regarded as related task. In previous research, Rich Caruna proposed that multi-task learning (MTL) can be implemented to fully leverage those related task [1]. Training with multiple related tasks simultaenously, classifier can learn generalized information compared to the classifier that classifier trained to classify only one variable. Also, during training with MTL, feature learnt from one task can be helpful to the other task. Rich Caruna, terminologied those positive effect as inductive transfer.

However, conventional studies do not leverage potential inductive transfer in fault classification. For example, if metadata of machine system has not only fault status but also operation condition and machine type, conventional researchers only focus on using fault status and neglect others. Also, if machine system has compound fault conditions, conventional studies trained classifier to classify those compound fault conditions as multi-class classification (MCC), instead of defining multiple classification task and fully leverage inductive transfer.

To overcome limitations of conventional studies, we trained a drone fault classifier to predict both fault condition (primary task) and maneuvering direction (auxiliary task) through MTL [2]. In this dataset, not only fault condition but also maneuvering direction effects signal measured from drone. During testing, we only evaluated fault predictions. The results revealed that the MTL-trained fault classifier outperformed the single-task learning (STL)-trained classifier. Thus, using MTL to classify both fault condition and other variable influencing measured signal enhances fault classification performance.

Furthermore, we adapted the cross-talk architecture to use MTL in fault classification [3]. Specifically, we proposed a novel cross-talk architecture with cross-talk layer (CTL) named residual neural dimension reductor (RNDR). We applied this architecture to compound fault classification for motors, where all severities of fault components influence signal measured from system. The model was trained to classify the severity of each fault component using MTL. Then, the predicted severity levels were aggregated to predict the compound fault status of motors. The results indicated that the proposed cross-talk architecture with RNDR outperformed not only STL methods but also conventional MTL methods based on the shared trunk architecture and other cross-talk architectures, sharing useful information through the CTL more effectively.

In this work, we examine whether the proposed cross-talk architecture outperforms conventional methods across various machine systems influenced by multiple variables. For this, we first used our drone fault dataset [2] to observe classification performance with not only shared trunk architecture we used before, but also cross-talk architecture. Second, we tested under the motor compound fault dataset [3]. In this paper, we tested without domain adaptation under partially labeled conditions, unlike the original paper. Instead, we tested the classification performance of classifiers in case the input is single-channel or multi-channel vibration data.

## 2. Multi-task learning models

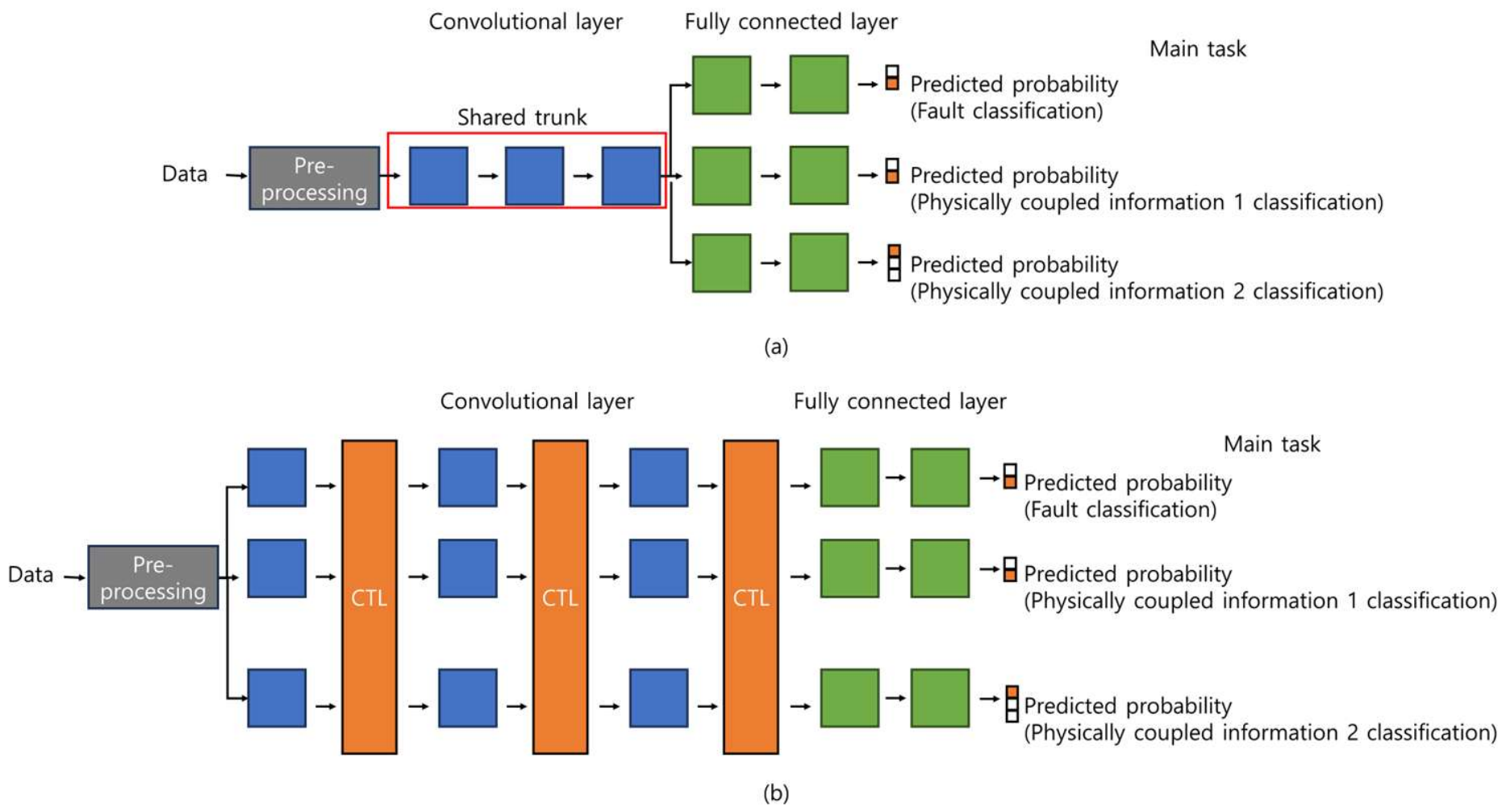

Figure 1: Multi-task learning models. (a) Shared trunk architecture, (b) Cross-talk architecture.

### 2.1 Shared trunk architecture

Shared trunk architectures are MTL models with a common feature extractor called a shared trunk, as illustrated in Fig. 1(a). We used the task-constrained deep convolutional network (TCDCN) [4] and the multi-task attention network (MTAN) [5]. TCDCN has a simple shared trunk and multiple task-specific branches for classification. Meanwhile, MTAN builds on the TCDCN structure by adding independent attention blocks for each task.

Through the shared trunk, the classifier learns general information, avoiding overfitting to specific fault patterns. This trait can be useful when common features are needed for each task consisting MTL. However, if the features needed to optimize both tasks are different, negative transfer can occur and degrade classification performance.

## 2.2 Cross-talk architecture

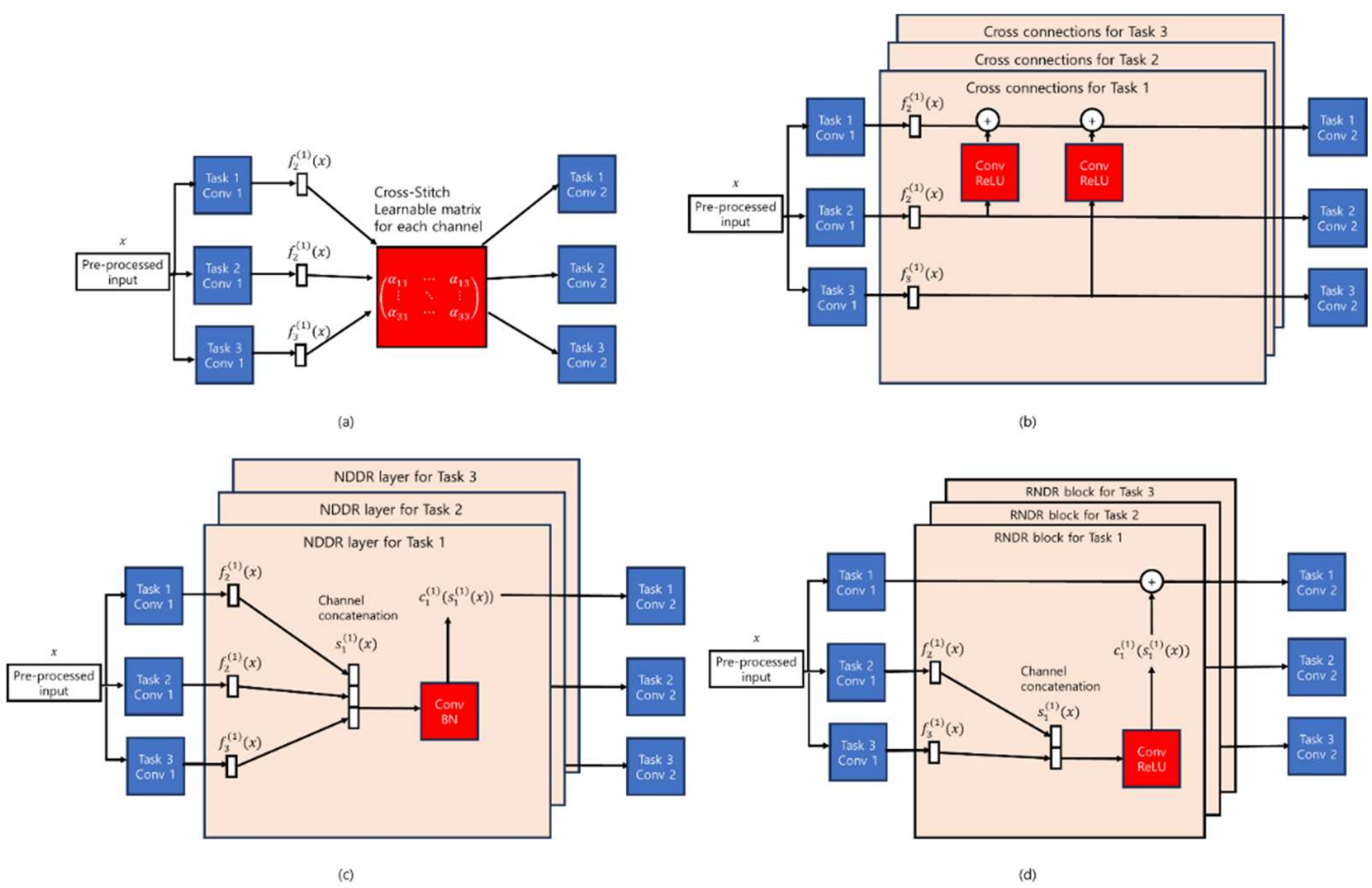

Figure 2: Cross-talk layers of cross-talk models. (a) Cross-stitch, (b) Cross-connected, (c) Neural-discriminative dimensionality reduction convolutional neural network, and (d) Residual neural dimension reductor.

Cross-talk architectures separate classifiers connected with CTLs from each other, as depicted in Fig. 1(b). Because these architectures have separate classifiers for each task, they can better concentrate on each task than shared trunk architectures. Moreover, each classifier obtains useful information from other classifiers through the CTLs. Therefore, unlike shared trunk architectures, cross-talk architectures can share information selectively, thereby leveraging the advantages of MTL while preventing negative transfer.

Specifically, we evaluated cross-stitch (CS) [6], cross-connected (CC) [7], neural-discriminative dimensionality reduction convolutional neural network (NDDR-CNN) [8], and our residual neural dimension reductor (RNDR) [3] as cross-talk architectures. Their CTL structures appear in Figs. 2(a)–(d), respectively. CS uses learnable matrices; CC uses convolutional layers with ReLU activation for information exchange. NDDR-CNN concatenates all the features in the channel dimension and processes the concatenated feature using its convolutional layer. The convolutional layer reduces the concatenated channel dimension to the original dimension. RNDR follows the NDDR-CNN structure but adds residual connections to preserve the original features.

## 3. Benchmarks for systems influenced by multiple variables

We prepared two benchmarks for systems influenced by multiple variables: the drone fault dataset [2] and the motor compound fault dataset [3]. The drone fault dataset includes stationary signals from drones operated at a constant speed (RPM). On the other hand, the motor compound fault dataset en-compasses various RPM operation conditions, including RPM changes in a sinusoidal pattern, a triangu-lar pattern, or a constant value. The goal of the drone fault dataset is to solely classify drone faults; therefore, we only use drone status for evaultation, even though we use other variables altogther as MTL. However, in the case of the motor compound fault dataset, we use every predicted fault component label in evaluation. To prevent confusion, we define two terms: the primary task is the classification task, which is also evaluated. The auxiliary task is the task that is not evaluated, only used for training and validation.

### 3.1 Drone fault dataset

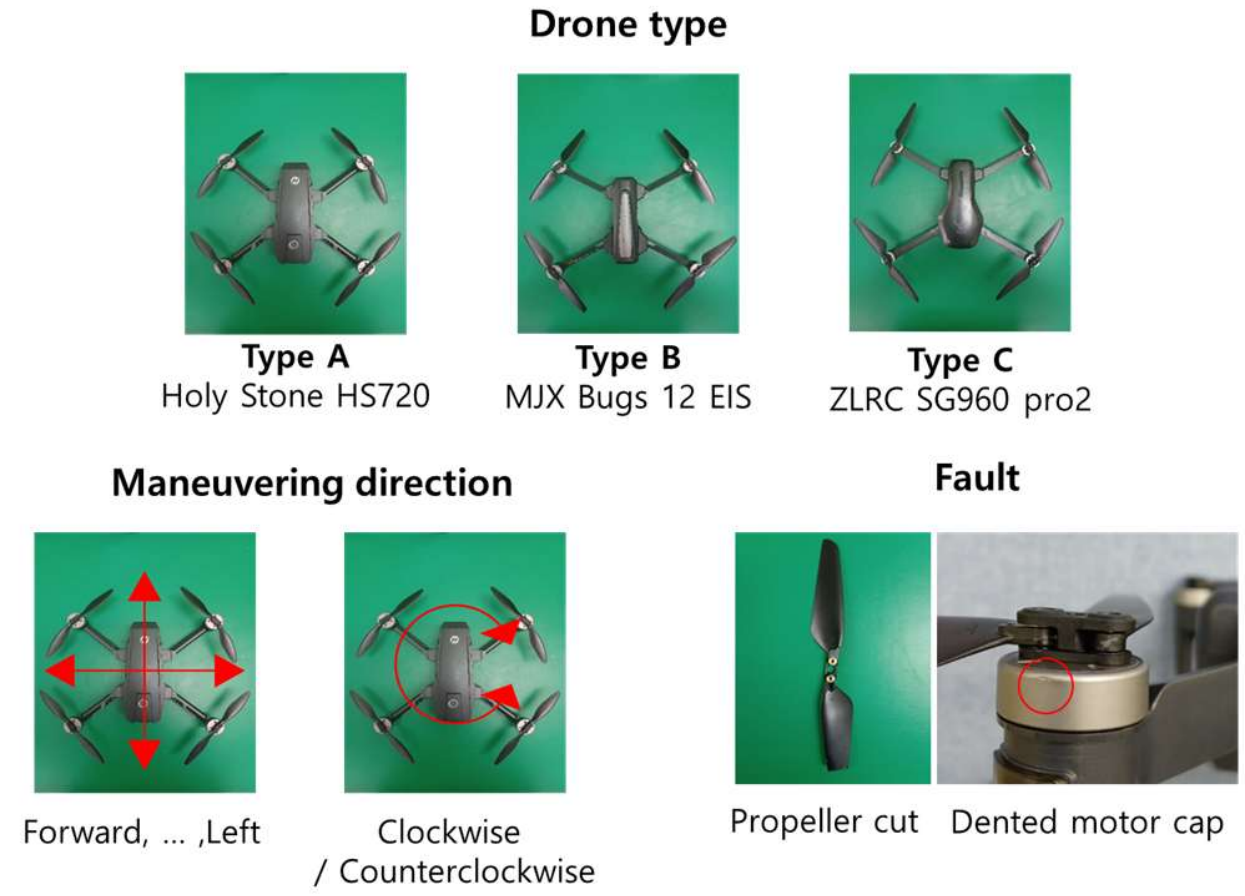

Figure 3: Drone type, maneuvering direction, and fault.

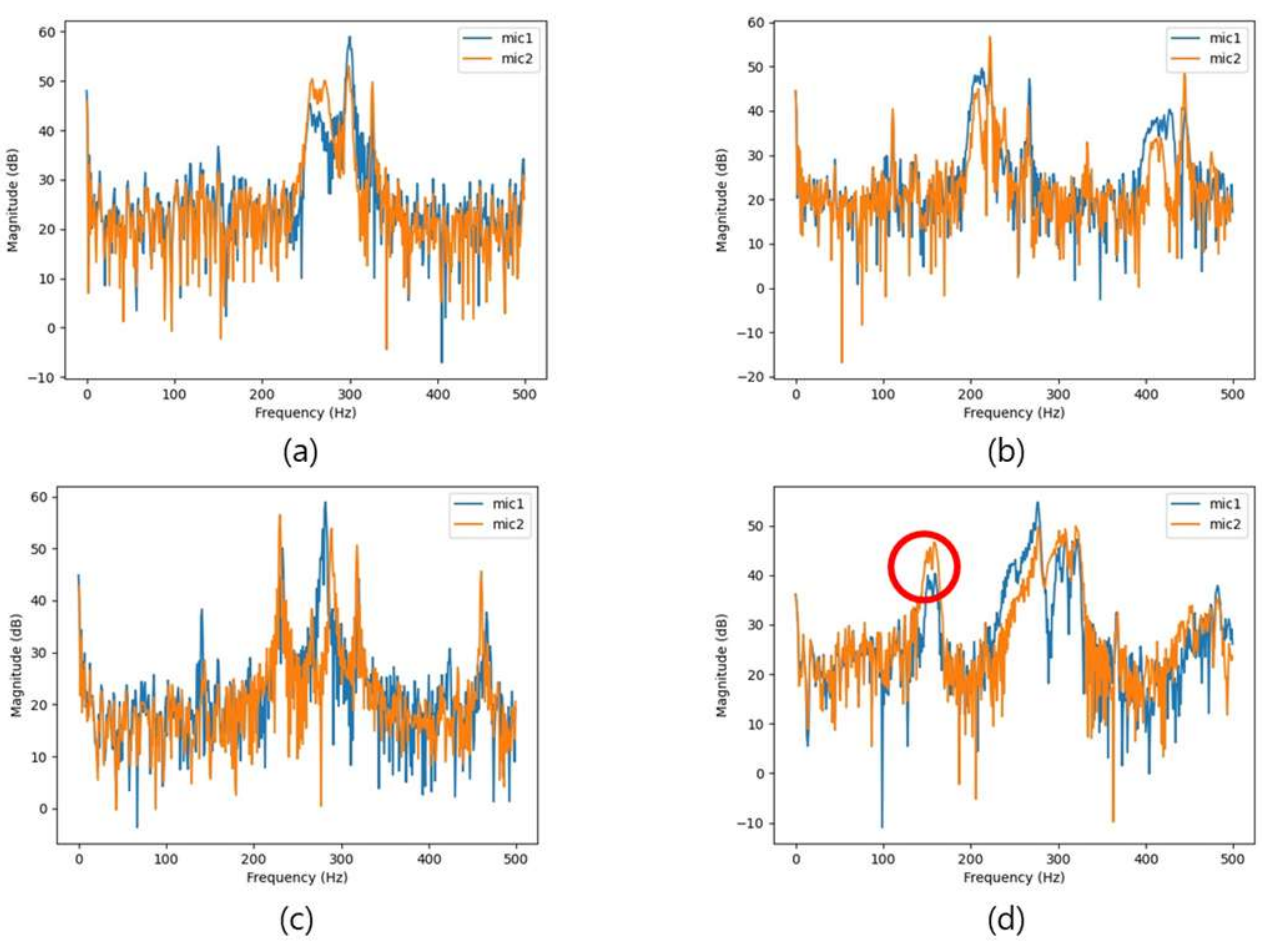

Figure 4: Power spectrum of drone sound.

The drone fault dataset contains fault, drone type, and maneuvering direction labels (Fig. 3). Drone type and maneuvering direction affect emitted sound, as shown by the power spectral differences in Fig. 4. Fig, 4(a)-(c) depict the power spectrum of normal drones. The signal in Fig. 4(a) was measured using a Holy Stone HS720 drone moving forward. The signal in Fig. 4(b) was measured using a ZLRC SG960

Pro 2 drone. The signals in these figures are distinct because they were measured using different drone models. Meanwhile, the signal in Fig. 4(c) was measured using a Holy Stone HS720 drone moving counterclockwise. This signal does not differ considerably from Fig. 4(a). Nevertheless, some differences exist, attributable to changes in maneuvering direction. We can leverage this property by using MTL. Fig. 4(d) depicts the power spectrum of a Holy Stone HS720 drone moving forward with a propeller cut.

Table 1: Congifugration of drone fault dataset.

| Primary task / Auxiliary task | Variable | Class | Number of classes | IBR |
|---|---|---|---|---|
| Primary task | Fault | Normal, Dented motor cap (4 locations), Propeller cut (4 locations) | 9 | 15.57 |
| Auxiliary task | Drone type | Holy Stone HS720, MJX Bugs 12 EIS, ZLRC SG960 Pro 2 | 3 | 1.03 |
| | Manuevering direction | Forward, Backward, Left, Right, Clockwise, Counterclockwise | 6 | 1.06 |

Details of the drone fault dataset are given in Table 1. Our dataset was subsampled from [1] to reflect the class-imbalance property resulting from a scarcity of fault conditions. To quantify the degree of class imbalance, we define the imbalance ratio (IBR) as the ratio between the largest and the smallest class size. The subsampled version can be downloaded from Zenodo (https://zenodo.org/records/17555246).

### 3.2 Motor compound fault dataset

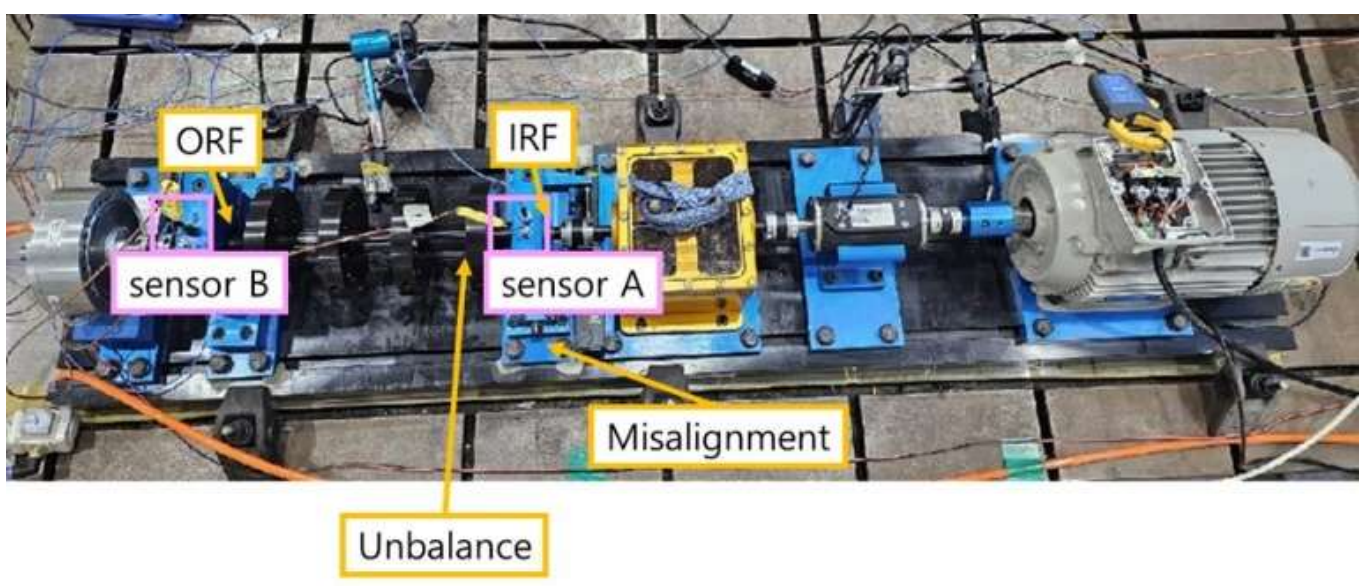

Figure 5: Motor testbed for motor compound fault.

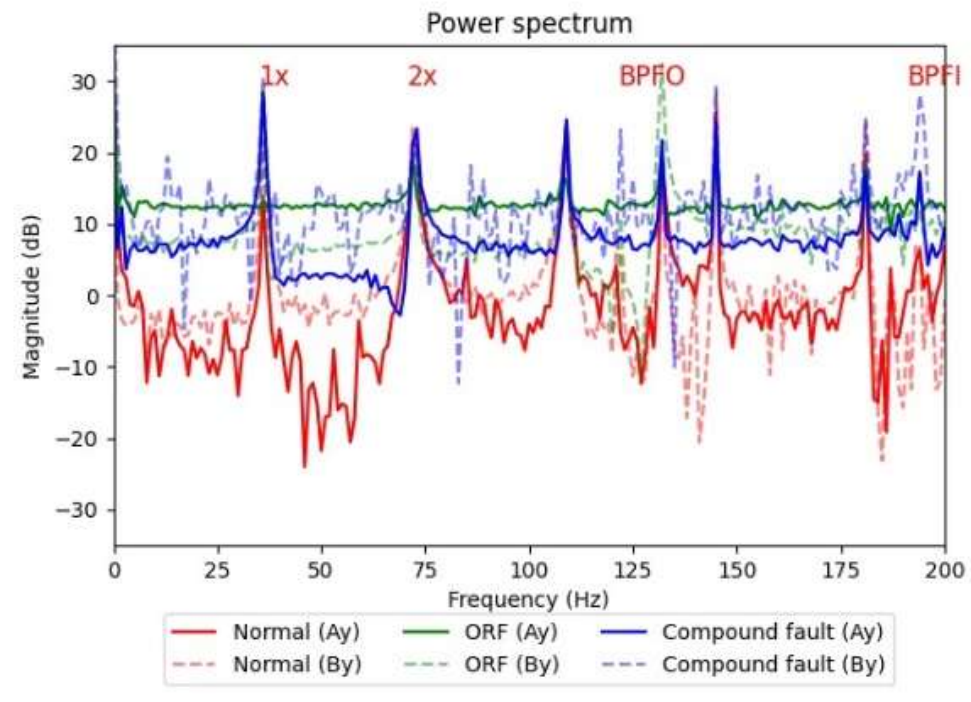

Figure 6: Power spectrum of vibration signal measured from sensor A and sensor B.

Table 2: Configurgation of the motor compound fault dataset.

| Primary task / Auxiliary task | Variable | Class | Number of classes | IBR |
|---|---|---|---|---|
| Primary task | Inner race fault | 0 mm, 0.2 mm | 2 | 1.68 |
| | Outer race fault | 0 mm, 0.2 mm | 2 | 1.68 |
| | Misalignment | 0 mm, 0.15 mm, 0.3 mm | 3 | 1.06 |
| | Unbalance | 0 g, 10 g, 18 g | 3 | 1.18 |

The motor compound fault dataset is measured from the motor testbed in Fig. 5. From the above Fig. 5, we induced four types of fault simultaneously: inner race fault (IRF), outer race fault (ORF), misalignment, and unbalance. When fault occurs simultaensously, spectral characteristic such as power spectrum is differed from single fault condition as Fig. 6. The detailed information is in Table 2. From Table 2, it may seem IBR is quite low. But when we aggregate each fault label to the compound fault label, IBR becomes 18.15; therefore, it reflects the scarcity of fault data. The dataset can be downloaded from Zenodo (https://zenodo.org/records/15743425), as subset E.

# 4. Experiment

## 4.1 Network architecture

We used short-time Fourier transform (STFT) for pre-processing. Then, we used a two-dimensional convolutional neural network (2D-CNN)-based backbone architecture. To the drone fault dataset, we applied frequency dynamic convolution [9] on the last two convolutional layers to account for the stationarity of the drone faults. For both two benchmarks, we used frequency layer normalization (FLN) for normalizing features in convolutinonal blocks [10]. With this setup, we compared the performance of the models trained using STL and MTL. For STL, we used only one classifier for the drone fault and we used four separate classifiers for each fault component, for motor copmound fault. As the MTL models, we compared TCDCN [4], MTAN [5], CS [6], CC [7], NDDR-CNN [8], and RNDR [3].

## 4.2 Training and validation

We used the categorical cross-entropy loss for each classification task, which together comprised the total loss function in our MTL framework. For training, we used the Adam optimizer [11] with a learning rate of 1e-3 over 100 epochs. For each epoch, we calculated the validation loss and selected the model with the lowest validation loss for the test.

## 4.3 Test

We used the macro F1 score as a performance metric to reflect the class-imbalance scenario. We tested each architecture three times with different random seeds, then averaged them for performance comparison. As mentioned above, we only use the fault label in the drone fault dataset; meanwhile, we use all labels of fault components in the motor compound fault dataset. Also, for the motor compound fault dataset, we tested performance with two cases: using sensor A only, and using both sensors, in order to see whether our MTL model works well for both single-input-multiple-output (SIMO) and multiple-input-multiple-output (MIMO) structures.

## 5. Result

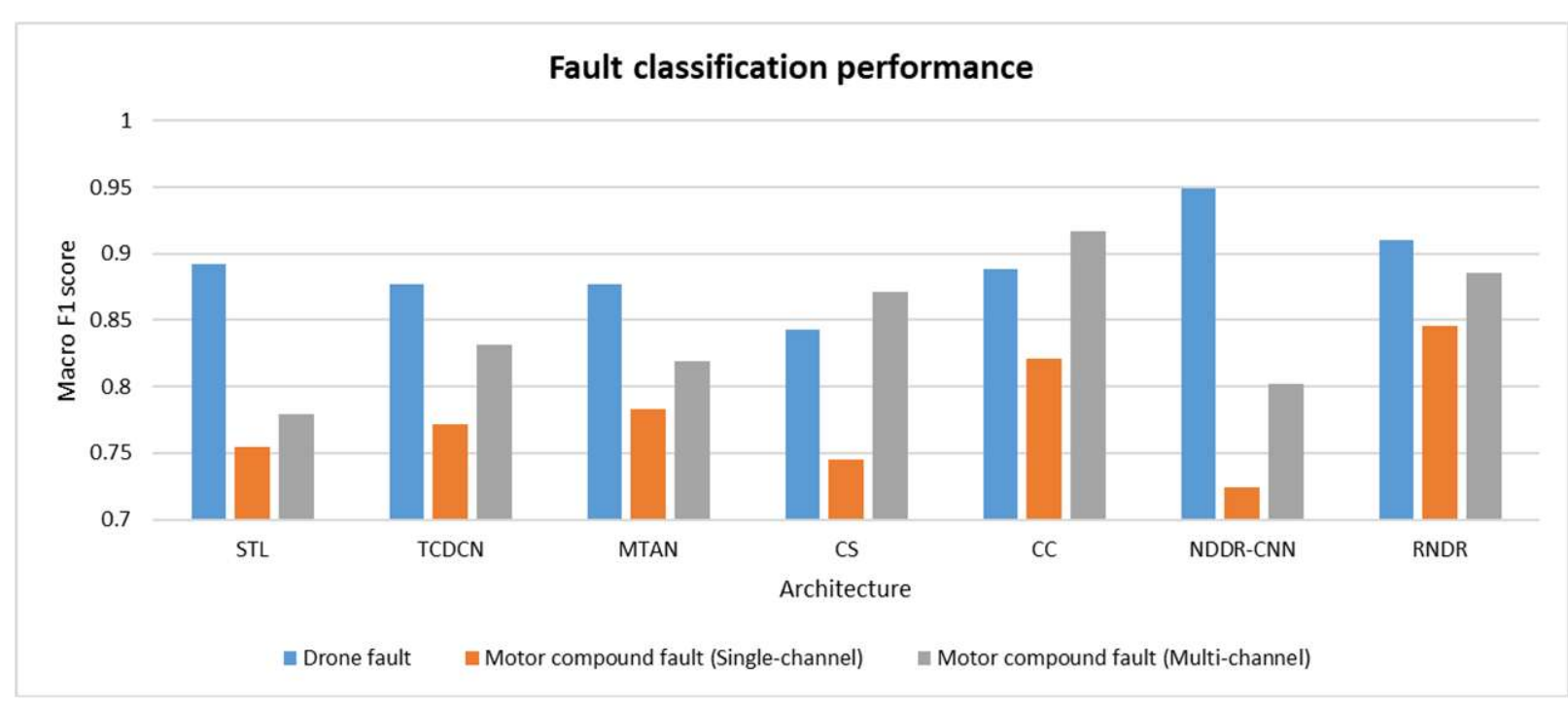

Figure 7: Fault classification performance.

Fig. 7 shows the fault classification performance of various architectures. Generally, the model trained with MTL showed better performance than the model trained with STL, except for some cases. To be specific, for drone fault, NDDR-CNN showed the highest classification performance, but RNDR still showed the highest classification performance. For the motor compound fault with single-channel data, RNDR outperformed other architectures. Lastly, for the motor compound fault with multi-channel data, CC has the best classification performance; still, RNDR showed the second-highest classification performance. In the case of CC and NDDR-CNN, they showed the best classification performance for certain benchmarks, but did not show good performance in other datasets. However, our RNDR showed the first or second rank in all benchmarks.

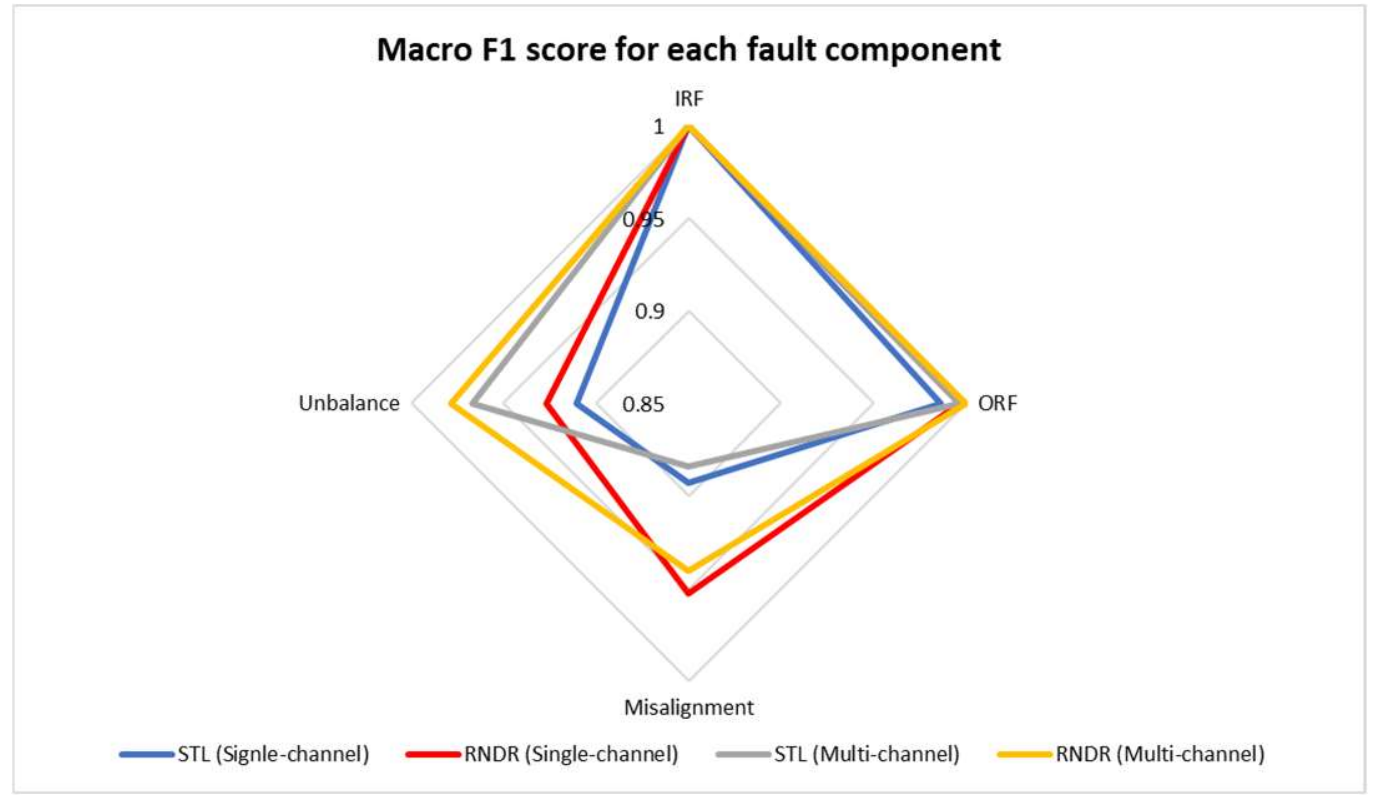

Figure 8: Macro F1 score for each fault component.

Fig. 8 shows the Macro F1 score of each fault component, instead of showing them in aggregated form. As Fig. 8 shows, from the model trained with STL, RNDR outperformed all the classification performances for each fault component. This indicates that model trained with MTL increased all fault severity classification performance of fault components using inductive transfer.

## 6. Conclusion

We showed that the MTL-trained classifiers outperformed the STL-trained classifier for fault classification of systems influenced by multiple variables. Additionally, we showed that among the MTL-trained classifiers, our proposed cross-talk architecture, RNDR, robustly outperformed other architectures. This was attributed to effective information exchange among the classifiers while mitigating negative transfer.

# Acknowledgements

This work was supported by the InnoCORE program of the Ministry of Science and ICT(N10260002).